\documentclass[letterpaper, 10 pt, conference]{ieeeconf}  

\IEEEoverridecommandlockouts                              

\usepackage{graphics} 
\usepackage{epsfig} 
\usepackage{amsmath} 
\usepackage{amssymb}  
\usepackage{multirow}

\title{\LARGE \bf
Adaptive Invariant Extended Kalman Filter for Legged Robot State Estimation
}

\author{Kyung-Hwan Kim, DongHyun Ahn, Dong-hyun Lee, JuYoung Yoon, and Dong Jin Hyun
\thanks{ This work was supported by the R\&D Division of Hyundai Motor Company. (\textit{Kyung-Hwan Kim is first author. Dong Hyun Ahn, Dong Hyun Lee, Ju Young Yoon are second authors.})(\textit{Corresponding authors: Dong Jin Hyun})
Kyung-Hwan Kim, Dong Hyun Ahn, Dong Hyun Lee, Ju Young Yoon, and Dong Jin Hyun are with the Robotics Lab, Research and Development Division, Hyundai Motor Company, Uiwang 16082, South Korea(e-mail: kkh9764@hyundai.com; dhahn@hyundai.com; lee.dh@hyundai.com; imax@hyundai.com; mecjin@hyundai.com)}}

\begin{document}

\maketitle
\begin{abstract}
 State estimation is crucial for legged robots as it directly affects control performance and locomotion stability. 
 In this paper, we propose an Adaptive Invariant Extended Kalman Filter to improve proprioceptive state estimation for legged robots. The proposed method adaptively adjusts the noise level of the contact foot model based on online covariance estimation, leading to improved state estimation under varying contact conditions. It effectively handles small slips that traditional slip rejection fails to address, as overly sensitive slip rejection settings risk causing filter divergence. Our approach employs a contact detection algorithm instead of contact sensors, reducing the reliance on additional hardware.
 The proposed method is validated through real-world experiments on the quadruped robot LeoQuad, demonstrating enhanced state estimation performance in dynamic locomotion scenarios.
\end{abstract}


\IEEEpeerreviewmaketitle
\section{Introduction}
The state of the legged robot, including body velocity and orientation, is crucial for legged robot control applications\cite{R.Grandia2023, F.Jenelten2022, J.DiCarlo2018, Ding2020RepresentationFreeMPC, Kim2019HighlyDQ, Li2024CafeMpcAC}. These states are challenging to measure directly and are typically estimated using sensor fusion methods \cite{M.Bloesh2012, M.Bloesh2013, M.Camurri2020, Park.Hae-WonMAP, Teng2021LeggedRS, Hartley2019ContactaidedIE, Rotella2014, Cheetah3, Y.Gao2022, Lin2021LeggedRS, S.Yang2023Cerverus}. The literature on legged robot estimation is classified based on the sensors used: \textit{proprioceptive}, \textit{exteroceptive}, or both\cite{M.Camurri2020}. Exteroceptive methods are useful but still face challenges in difficult environments such as fog or reed fields. In this paper, our objective is to improve the performance of proprioceptive estimation for control.

Bloesch et al. \cite{M.Bloesh2012} introduced a proprioceptive method that fuses leg kinematics, contact data, and inertial measurements (accelerometer and gyroscope) using an Extended Kalman Filter (EKF) with static contact assumption. 
However, it is vulnerable to slipping due to the static contact assumption. In the following paper \cite{M.Bloesh2013}, they proposed a model using static contact velocity kinematics with UKF for higher accuracy and introduced Mahalanobis distance based slip rejection for slippery terrain.  

A similar methodology has been applied to bipedal robots\cite{Teng2021LeggedRS}\cite{Hartley2019ContactaidedIE}\cite{Rotella2014}. Building on the above idea, R. Hartley et al. \cite{Hartley2019ContactaidedIE} implemented the Invariant Extended Kalman Filter (IEKF) \cite{Bonnabel2017}, which offers faster convergence and improved robustness than EKF. Despite the unobservability of the absolute position, it demonstrated low drift in long odometry experiments. However,  it is also vulnerable to slip due to the static contact assumption and it does not address the slip. Teng et al. \cite{Teng2021LeggedRS} proposed a methodology that utilizes IEKF and Mahalanobis distance-based slip rejection, similar to \cite{M.Bloesh2013}. 

Bledt et al. \cite{Cheetah3} employed a two-stage approach, first estimating orientation and then estimating position and velocity using a Kalman Filter (KF) with a contact detection algorithm. This method achieved fast estimation, but did not address slip and has the limitation that contact and kinematics cannot influence orientation estimation. 

Kim, J.H. et al. \cite{Park.Hae-WonMAP} proposed an estimation method using Maximum A Posteriori (MAP) in a similar model. They also suggested slip rejection by predicting the current foot velocity from previous velocity estimates. However, the large size of the estimator limits implementation in real-time.


We categorize previous models into two groups: the Position Kinematics-based Model (PKM) and the Velocity Kinematics-based Model (VKM). PKM uses the foot position, assumed to be static, as a state variable and uses position kinematics as a measurement \cite{M.Bloesh2012, Hartley2019ContactaidedIE}. Otherwise, VKM uses velocity kinematics measurement, assuming the foot is static \cite{M.Bloesh2013, Teng2021LeggedRS}. While VKM features a smaller filter size and is more effective at handling slips, it exhibits higher drift compared to PKM due to the absence of landmark positions and noise in velocity measurement. To address this, we adopt PKM and incorporate slip detection. 

Previous studies on PKM typically used a small constant value for the contact foot model noise during the stance phase. However, since this approach is vulnerable to slips, slip rejection methodologies are typically added. Two main slip rejection methodologies are commonly used: one based on the Mahalanobis distance \cite{M.Bloesh2013} and the other predicting the current foot velocity \cite{Park.Hae-WonMAP}. These methods rely on a threshold-based on/off approach, which makes it difficult to handle small slips or address individual x, y, and z axes. If the slip rejection is set too sensitively, all legs may be classified as slipping, causing the filter to diverge. To address this issue, we propose a method that adaptively adjusts the contact foot model noise.
Furthermore, in contrast to most previous studies that rely on contact sensors, the proposed approach is validated using a contact detection algorithm\cite{Bledt2018Contact}, which reduces hardware costs. We developed the methodology based on the IEKF, ensuring robustness and fast convergence. The proposed method was tested on our quadruped robot, LeoQuad.

The contributions of this work are as follows:
\begin{itemize}
    \item Proposed an Adaptive IEKF that dynamically adjusts the noise level of the contact foot model.
    \item Integrated a Mahalanobis distance-based slip rejection method for the PKM.
    \item Validated the proposed methodology through real-robot experiments with a contact detection algorithm.
\end{itemize}



Section \uppercase\expandafter{\romannumeral2} provides the mathematical overview that is essential to be understood in advance. 
Section  \uppercase\expandafter{\romannumeral3} discusses the state estimation methodology for legged robots.
Section  \uppercase\expandafter{\romannumeral4} presents the results from real-robot experiments.
Finally, we conclude in Section \uppercase\expandafter{\romannumeral5}.


\section{Prerequisites}
We briefly introduce the mathematical prerequisites used in this study: $SE_{2+N}$(3) group and the Right-Invariant Extended Kalman Filter (RIEKF).

\subsection{$SE_{2+N}(3)$ Group Representation}
Let $G$ be a matrix Lie group and $\mathfrak{g}$ its corresponding Lie algebra. One of the matrix Lie group, $SE_{2+N}$(3), represents rigid body motion in 3D space with position of $N$-landmarks:
\begin{equation}
\resizebox{0.43\textwidth}{!}{$
SE_{2+N}(3) = 
\left\{
\chi = \begin{pmatrix}
    R & v & p & r_1 & \cdots & r_N \\
    \boldsymbol{0}_{1 , 3} & 1 & 0 & 0 & \cdots & 0 \\
    \boldsymbol{0}_{1 , 3} & 0 & 1 & 0 & \cdots & 0 \\
    \boldsymbol{0}_{1 , 3} & 0 & 0 & 1 & \cdots & 0 \\
    \vdots & \vdots & \vdots & \vdots & \ddots & \vdots \\
    \boldsymbol{0}_{1 , 3} & 0 & 0 & 0 & \cdots & 1
\end{pmatrix}, \,
R \in SO(3), \, v, p, r_i \in \mathbb{R}^3
\right\}$}
\end{equation}
The $R$, $v$, and $p$ are a rigid body's rotation, velocity, and position, respectively. The $r_1$ to $r_N$ are each landmarks. The corresponding Lie algebra $\mathfrak{se}_{2+N}(3)$ is defined as follow:
\begin{equation}
\resizebox{0.43\textwidth}{!}{$
   \mathfrak{se}_{2+N}(3) =
   \begin{Bmatrix}
       \xi^\wedge=\begin{pmatrix}
           \omega^\times & a & v & u_1 & \cdots & u_N \\
           \boldsymbol{0}_{1, 3} & 0 & 0 & 0 & \cdots & 0 \\
           \vdots & \vdots & \vdots & \vdots & \ddots & \vdots \\
           \boldsymbol{0}_{1, 3} & 0 & 0 & 0 & \cdots & 0
       \end{pmatrix}, \,
      \xi = \begin{pmatrix}\xi_R\\\xi_v \\\xi_p \\\xi_{r_1} \\\vdots\\\xi_{r_N} \end{pmatrix}\in \mathbb{R}^{9+3N}
   \end{Bmatrix}
$}
\end{equation}
The operator $(\cdot)^\wedge:\mathbb{R}^{dim \mathfrak{g}}\rightarrow \mathfrak{g}$ is used to map vector space elements to tangent space. The operator $(\cdot)^\times: \mathbb{R}^3\rightarrow \mathfrak{so}(3)$ is one of the $(\cdot)^\wedge$ operator in $SO$(3) groups. It is defined so that $[x]^{\times}y=x\times y$. The $\omega$ and $a$ are angular velocity and acceleration of the rigid body, respectively, and $u$ is the velocity of landmarks.
Lie groups and their corresponding Lie algebras are related through the exponential map:
\begin{align}
    \exp(\cdot)&:\xi^\wedge\rightarrow\chi\\
    \text{Exp}(\cdot)&:\xi\rightarrow\chi
\end{align}
The \textit{adjoint} matrix at $\chi$, $Ad_{\chi}$, describes how a Lie algebra elements transforms under the group action.
\begin{equation}
    (Ad_{\chi}\xi)^\wedge = \chi \xi^\wedge \chi^{-1}
\end{equation}
For more details, refer to \cite{Bonnabel2017}\cite{Chirikjian2001}

\begin{figure*}[ht]
    \centering
    \includegraphics[width=0.95 \textwidth]{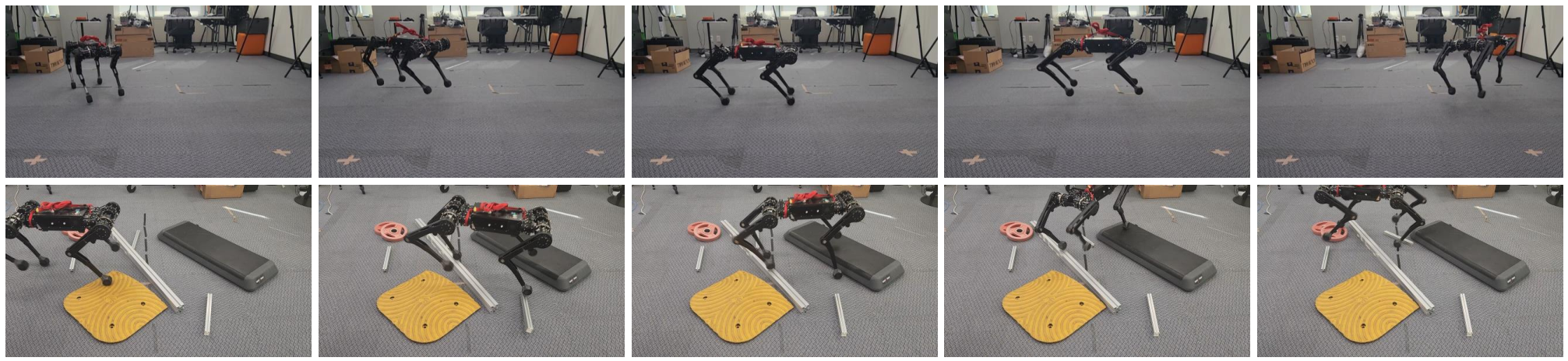} 
    \caption{The snapshot of robot, LeoQuad, experiment. The top row is a snapshot of a pronk on flat terrain, while the bottom is a snapshot of a flying trot on rough terrain.}
    \label{fig:real_robot} 
\end{figure*}

\subsection{Right Invariant Extended Kalman Filter}
The Invariant Extended Kalman Filter (IEKF), introduced in \cite{Bonnabel2007}\cite{Bonnabel2017}, is a variant of the EKF that exploits the Lie groups structure. The IEKF computes the gain independently of the estimated state. This characteristic allows the IEKF to maintain local stability properties, which the EKF does not possess\cite{Bonnabel2017}.
The general form of the stochastic dynamics system and right-invariant measurement is given as follows:
\begin{subequations}\label{eq:RIEKF_model}\begin{align}
    \dot{\chi}_t&=f(\chi_t,u_t) + \chi_t w_t^{\wedge}\label{eq:inekf_dynamics}\\
      Y_k&=\chi^{-1}_k b_k +V_k \label{eq:inekf_measurement}
\end{align}\end{subequations}
The $k$ is discrete-time, such that $t=kT$, where $T$ is the sampling period. The $f$ is the continuous-time dynamics. The $u$ is an input and the $w$ is a continuous white Gaussian noise with a covariance matrix $Q_k$. The $b$ is a known vector and $V_k$ is the vector that contains the white Gaussian noise $v_k$ with a covariance matrix $N_k$. 
The RIEKF is defined as:
\begin{subequations}\label{eq:RIEKF}
\begin{align}
    \hat{\chi}_{k}^- &= f^d(\hat{\chi}_{k-1}^+,u_k) \text{, Propagation}\label{eq:propagation} \\ 
    \hat{\chi}^+_k &= \text{Exp}\left(K_k\begin{pmatrix}
       s^1(\hat{\chi}_k Y_k^1 - b_k^1)\\
       \vdots \\
       s^N(\hat{\chi}_k Y_k^N - b_k^N)
    \end{pmatrix}\right)\hat{\chi}_{k}^- \text{, Update} \label{eq:update}
\end{align}
\end{subequations}
The superscript $(\cdot)^-$ denotes the predicted (prior) value from the dynamics system, while $(\cdot)^+$ denotes the posterior value after incorporating the residual from the measurement model. The superscript $\hat{(\cdot)}$ denotes the estimated value, distinguishing it from the true value. The function $f^d$ is the discretization of $f$. The $s^i$ is a selection matrix for the $i$-th measurement model that appropriately reduces unnecessary dimensions. The matrix $K_k$ is an optimal gain matrix. The gain $K_k$ is computed through the following procedure. 

Consider two distinct trajectories $\chi_t$ and $\hat{\chi}_t$. The right-invariant error is defined as 
\begin{equation}\label{eq:invariant_error}
    \eta^R_t = \hat{\chi}_t \chi^{-1}_t \in G
\end{equation}
If the deterministic dynamics in Eq. \eqref{eq:inekf_dynamics} satisfy \textit{group-affine} condition\cite{Hartley2019ContactaidedIE}
\begin{equation}\label{eq:group_affine}
    f(\chi_1 \chi_2 ,u_t ) = f(\chi_1 ,u_t)\chi_2 +\chi_1 f(\chi_2 ,u_t )-\chi_1 f(I_d,u_t )\chi_2 ,
\end{equation}
the right-invariant error $\eta^R_t$ have a \textit{state-trajectory independence propagation}\cite{Bonnabel2017}, which can be expressed differential equation as follow:
\begin{equation}\label{eq:error_dynamics_eta}
    \frac{d}{dt}\eta_t^R=g(\eta_t^R,u_t)-(Ad_\chi w_t)^\wedge\eta_t^R ,
\end{equation}
where, $g(\eta_t^R,u_t)=f(\eta_t^R,u_t)-\eta_t^R f(I_d,u_t)$. If the error is assumed to be small, we can define a vector error variable $\xi_t$ as 
\begin{equation}\label{eq:log_linear}
    \eta^R_t = \exp(\xi_t^\wedge)
\end{equation}
From the Eq. \eqref{eq:error_dynamics_eta} and the first-order approximation $\exp(\xi_t^\wedge)\approx I+\xi^\wedge$, we can find a linear approximation of the error dynamics
\begin{equation}\label{eq:error_propagation}
    \frac{d}{dt}\xi_t = A_{u_t} \xi_t - Ad_{\hat{\chi}}w_t
\end{equation}
The propagation matrix $A_{u_t}$ is defined by $g(\exp(\xi_t^\wedge),u_t) = (A_{u_t}\xi)^{\wedge}+O(\|\xi_t\|^2)$. 

From Eq. \eqref{eq:update}, \eqref{eq:invariant_error}, and \eqref{eq:log_linear}, the linearized error update is as follows:
\begin{equation}\label{eq:error_update}\scalebox{0.8}{$
    \xi^+_k = \xi_k + K_k
    \begin{pmatrix}
        s^1(\xi_k^{\wedge}b_k^1 +\hat{\chi}_k V_k^1)\\
        \vdots\\
        s^m(\xi_k^{\wedge}b_k^m +\hat{\chi}_k V_k^m)
    \end{pmatrix}\\
    =\xi_k - K_k(H_k\xi_k - M_kv_k)$}
\end{equation}
$H_k$ is a measurement matrix constructed by stacking smaller matrices $H_k^1, H_k^2, \dots, H_k^m$ vertically, where each block $H_k^i$ is defined as $H_k^i \xi_k^i = -s^i \xi_k^\wedge b_k^i$. $M_k$ is a block diagonal matrix composed of smaller matrices $M_k^1, M_k^2, \dots, M_k^m$, where each block $M_k^i$ is defined as $M_k^i v_k^i = s^i \chi V_k^i$. The gain $K_k$ is computed using standard discrete-time Kalman Filter equations\cite{OptimalStateEstimation}: 
\begin{subequations}\small\label{eq:kalman_filter_gain}\begin{align}
P_k^- &= A_{k-1}^d P_{k-1}^+{A_{k-1}^d}^T+ Ad_{\hat{\chi}^-} Q_k^d Ad_{\hat{\chi}^-}^T\label{eq:p_prediction}\\ 
S_k &= H_k P_k^- H_k^T + M_k N_k M_k^T \\
K_k &= P_k^- H_k^T S_k^{-1} \\
P_k^+ &= (I-K_k H_k)P_k^-
\end{align}
\end{subequations}
The $A_{k-1}^d$ is the discretization of $A_{u_t}$. The discretized covariance $Q_k^d$ can be approximated as $Q_k^d\approx Q_kT$\cite{OptimalStateEstimation}.

\section{Methodology}
We represent the states and biases to be estimated as: 
\begin{subequations}\small
\begin{align}
   \chi&=
    \begin{pmatrix}
        R_B & v_B & p_B & p_{F_1}& \cdots&p_{F_N}\\
        \boldsymbol{0}_{1, 3} &1 &0&0&\cdots&0\\
        \boldsymbol{0}_{1, 3} &0 &1&0&\cdots&0\\
        \boldsymbol{0}_{1, 3} &0 &0&1&\cdots&0\\
        \vdots & \vdots & \vdots& \vdots&\ddots& \vdots\\
        \boldsymbol{0}_{1, 3} &0 &0&0&\cdots&1\\
    \end{pmatrix}\\
b &=\begin{pmatrix}b_a&b_\omega\end{pmatrix}
    \end{align}
\end{subequations}
The superscripts $B$ and $F_i$ represent the body and $i$-th foot frame coordinates, respectively.
$p_{F_i}$ denotes the $i$-th foot position. The $N$ is the number of legs. The $b_a$ and $b_w$ denote the bias of the accelerometer and gyro in the body frame. The states are expressed in the inertial frame. For readability, the time $t$ and sampling time $k$ are omitted.

\subsection{System Dynamics \& Measurement Model}
The system dynamics of legged robots is generally formulated as described in \cite{M.Bloesh2012} and \cite{Hartley2019ContactaidedIE}:
\begin{subequations}\label{eq:system_dynamics}
\begin{align}
\dot{R}_B&=R_B(\omega-w_\omega)^{\times}\\
\dot{v}_B&=R_B(a-w_a)+g\\
\dot{p}_B&=v_B\\
\dot{p}_{F_i}& = R_B (-w_{f_i})\label{eq:foot_model}\\
\dot{b}_\omega &= w_{b\omega}\\
\dot{b}_a &= w_{ba}
\end{align}
\end{subequations}
\begin{equation}
     \begin{pmatrix}
         \omega\\
         a
    \end{pmatrix}=
    \begin{pmatrix}
        \tilde{\omega}-b_{\omega}\\
         \tilde{a}-b_{a}\\
    \end{pmatrix}
\end{equation}
The $\omega$ and $a$ are bias-corrected angular velocity and acceleration. The $\tilde{\omega}$ and $\tilde{a}$ are measured values by the gyroscope and accelerometer, which are corrupted by sensor noise denoted as $w_\omega$, $w_a$. The $b_\omega$ and $b_a$ are the corresponding biases with noise $w_{b\omega}$, $w_{ba}$. The $g$ is the gravitational acceleration vector. $w_{f_i}$ is each $i$-th foot linear velocity noise.  The noise terms are assumed to be white Gaussian noise with constant variance. The $w_{f_i}$ is set to be small during contact and larger during swing phases, thereby relaxing the static assumption of the foot model \eqref{eq:foot_model}.
System dynamics can be formed as an invariant observer \eqref{eq:inekf_dynamics}, and it satisfies Eq. \eqref{eq:group_affine}. 
\begin{subequations}\small
\begin{align}
    \dot{\boldsymbol{\chi}}_t&=f(\chi_t,u_t)-\chi_t w^{\wedge}\\
    f&=
    \begin{pmatrix}
        R_{B}\omega^\times & R_{B}a+g & v_B & \boldsymbol{0}_{3, 1}& \cdots& \boldsymbol{0}_{3, 1}\\
        \boldsymbol{0}_{1, 3} &0 &0  & 0& \cdots&0\\
        \vdots & \vdots & \vdots & \vdots&\ddots &\vdots\\
          \boldsymbol{0}_{1, 3} & 0 &0  &0 & \cdots &0\\
    \end{pmatrix}\\
    w^{\wedge}&=
     \begin{pmatrix}
        {w_\omega}^\times & w_a & \boldsymbol{0}_{3,1} & w_{f_1}& \cdots&w_{f_N}\\
        \boldsymbol{0}_{1, 3} &0 & 0 &0 &\cdots &0\\
        \vdots & \vdots & \vdots & \vdots & \ddots &\vdots\\
          \boldsymbol{0}_{1, 3} & 0 &0  &0 &\cdots &0\\
          \end{pmatrix}
\end{align}
\end{subequations}
The Right-Invariant error and the bias error are given by:
\begin{subequations}
\begin{align}
\eta^r &= \hat{\chi} \chi^{-1}\\
    \epsilon_b &= \begin{pmatrix}
        \epsilon_{b\omega}\\
       \epsilon_{ba}
    \end{pmatrix} = \begin{pmatrix}
        b_\omega-\hat{b}_\omega \\
        b_a- \hat{b}_a  
    \end{pmatrix}
\end{align}
\end{subequations}
The linear error dynamics are derived as follows:
\begin{subequations}\small
\begin{align}
  &\frac{d}{dt}\begin{pmatrix}
      \xi\\\epsilon_b
  \end{pmatrix} =A\begin{pmatrix}
      \xi\\\epsilon_b
  \end{pmatrix}+\begin{pmatrix}
     Ad_{\hat{\chi}} &\bf{0}\\
     \bf{0}&I
  \end{pmatrix}w\\
  &A=\begin{pmatrix}
     \bf{0} & \bf{0} & \bf{0} & \bf{0}&\cdots& \bf{0} &-R_B& \bf{0} \\
     [g]^{\times} & \bf{0} & \bf{0} & \bf{0}&\cdots& \bf{0} &-[v_B]^{\times}R_B& -R_B \\
     \bf{0} & I & \bf{0} & \bf{0}&\cdots& \bf{0} &-[p_B]^{\times}R_B& \bf{0} \\
     \bf{0} & \bf{0} & \bf{0} & \bf{0}&\cdots& \bf{0} &-[p_{F_1}]^{\times}R_B& \bf{0} \\
     \bf{0} & \bf{0} & \bf{0} & \bf{0}&\vdots& \bf{0} &\vdots& \bf{0} \\
     \bf{0} & \bf{0} & \bf{0} & \bf{0}&\cdots& \bf{0} &-[p_{F_N}]^{\times}R_B& \bf{0}\\
    \bf{0} & \bf{0} & \bf{0} & \bf{0}&\cdots& \bf{0} & \bf{0}& \bf{0}\\
    \bf{0} & \bf{0} & \bf{0} & \bf{0}&\cdots& \bf{0} &\bf{0}& \bf{0}
    \end{pmatrix}\\
    &Q_k = \mathbb{E}[ww^T] 
\end{align}
\end{subequations}
The bold zero is a 3 by 3 matrix.

The measurement model of leg kinematics is given as follows: 
\begin{equation}
    {}^{B}\tilde{p}_{F_i/B} = R_B^T(p_{F_i} - p_{B}) + w_p 
\end{equation}
${}^{B}\tilde{p}_{F_i/B}$ is the local forward kinematics computed from joint encoder measurements. The $w_p$ is the noise of the model. The noise may include the encoder noise\cite{Hartley2019ContactaidedIE}.
The above models can be expressed as Right-invariant observation form \eqref{eq:inekf_measurement} as follows:
\begin{equation}\label{eq:right_invariant_kinematics_model}
    \scalebox{0.9}{$
\begin{aligned}
       Y_{p_i}&=
        \begin{pmatrix}
        {}^{B}\tilde{p}_{F_i/B}\\0\\1\\-\delta_{1i}\\ \vdots\\-\delta_{Ni}
        \end{pmatrix}=
    \chi^{-1}
    \begin{pmatrix}
        \bf{0}_{3, 1}\\0\\1\\-\delta_{1i}\\ \vdots\\-\delta_{Ni}
    \end{pmatrix}+
    \begin{pmatrix}
        w_{p.i}\\0\\\vdots\\0
    \end{pmatrix},\\
        \delta_{ni} &= \begin{cases}
        1& \text{if } n=i, \\
        0& \text{if } n \neq i, \text{ where, } i=1,2,...,N\\
    \end{cases}
\end{aligned}$}
\end{equation}
The function $\delta$ is used to represent the kinematics observation model for multiple legs in a single equation. $w_{p.i}$ is the noise in the kinematics model of the $i$-th leg. From Eq. \eqref{eq:error_update},  the matrix for the Kalman filter update can be derived as follows:
\begin{subequations}\small
\begin{align}
    H_{p_i} &= \begin{pmatrix}
        \bf{0}_{3,3}& \bf{0}_{3,3} & -I_{3,3} &\delta_{1i}I_{3,3}& \cdots&\delta_{Ni}I_{3,3}
    \end{pmatrix}\\
    M_{p_i} &= R_B \\
    N_{p_i} &= \mathbb{E}[w_{p_i}w_{p_i}^T] 
\end{align}
\end{subequations}
Based on the system model and measurement model matrices derived in this way, the state is estimated through \eqref{eq:RIEKF} and \eqref{eq:kalman_filter_gain}.
\begin{figure}[htbp]
    \centering
    \includegraphics[width=0.475 \textwidth]{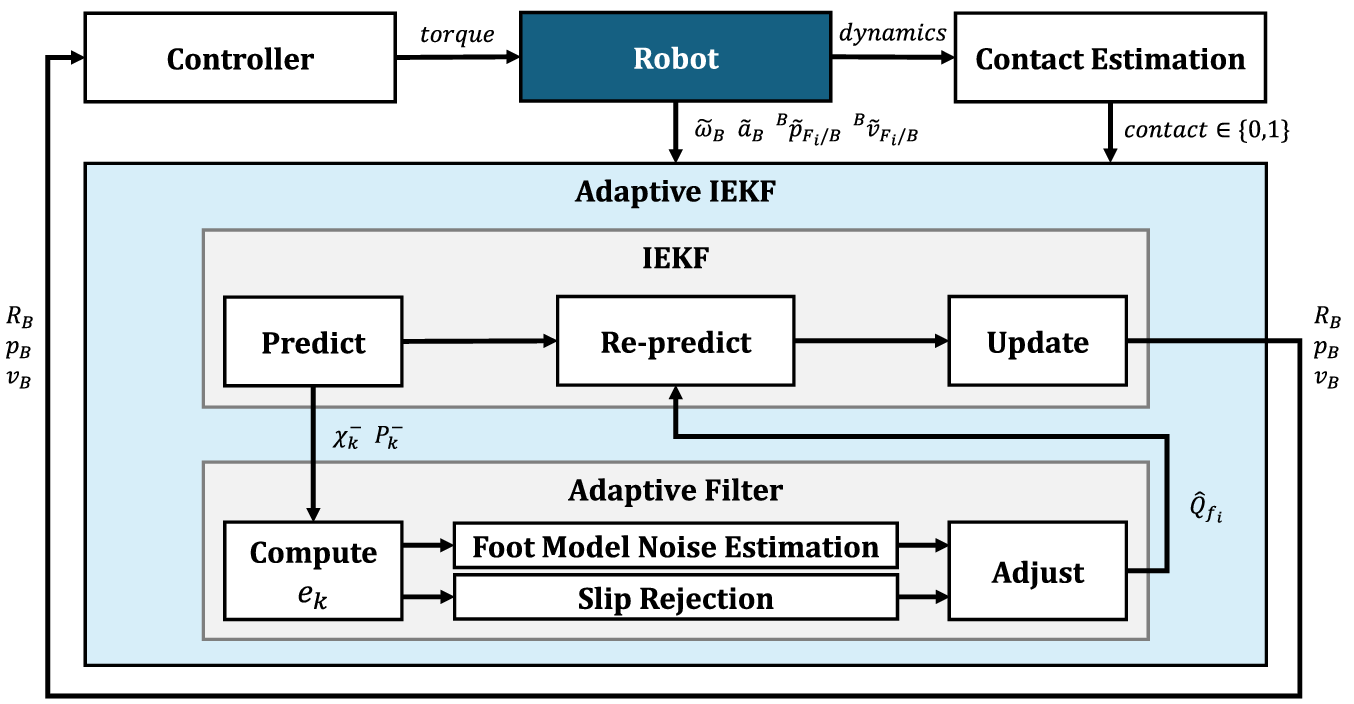} 
    \caption{Flowchart of Adaptive IEKF algorithm.}
    \label{fig:algorithm_flow_chart} 
\end{figure}

\subsection{Contact Detection Algorithm}
One key distinction in this study is the use of contact data obtained through contact detection algorithms rather than contact sensors. The contact detection is free from the costs or failures associated with contact sensors. Although it may lead to misdetection, these can be addressed by adaptive filters, which will be explained later.
Since contact detection is not a new method, we only briefly discuss it. Similar to \cite{Bledt2018Contact}, we fuse gait schedules with disturbance forces estimated via General Momentum-based (GM) methods using a Kalman Filter. Additionally, we used commanded torques in GM-based estimation instead of joint torques, as no torque sensors were employed.

\subsection{Adaptive Filtering - Foot Model Noise Estimation}
We used the velocity-kinematic model to estimate the noise of the contact foot state. The velocity kinematics model is given as follows:
\begin{equation}\label{eq:velocity_kinematics}
\dot{p}_{F_i} =v_B + R_B\omega^{\times}{}^{B}\tilde{p}_{F_i/B} + R_B{}^{B}\tilde{v}_{F_i/B} -R_B w_v
\end{equation}
The $w_v$ is a compound noise from multiple sources, including gyro and encoder noise. Since modeling this noise precisely is difficult, a reasonable constant value is manually chosen by the user to reflect their combined effect. 

Reconsidering the foot model \eqref{eq:foot_model}, the actual foot model behavior under slip or missed contact includes an unknown foot model noise $w_{uf_i}$, which is larger than the static assumed foot noise $w_{f_i}$. In such case, the model can be represented as $\dot{p}_{F_i} = R_B (-w_{uf_i})$. Substituting this equation into \eqref{eq:velocity_kinematics} yields the following equation:
\begin{equation}
   - \omega^{\times}{}^{B}\tilde{p}_{F_i/B} - {}^{B}\tilde{v}_{F_i/B}
    = R_B^T v_B + w_{uf_i} +w_v
\end{equation}
This can be expressed as a Right-invariant measurement model:
\begin{equation}\small\label{eq:vk_right_invariant_measurement}
\begin{aligned}
        &Y_{v_i} = X^{-1} b_{v_i} + V_{v_i}, \\
        &Y_{v_i} = \begin{pmatrix} - \omega^{\times}{}^{B}\tilde{p}_{F_i/B} - {}^{B}\tilde{v}_{F_i/B} \\ 0 \\ \vdots\\0 \end{pmatrix}, \\
        &b_{v} = \begin{pmatrix} \bf{0_{1.3}} & -1 & 0 & \cdots & 0 \end{pmatrix}^T,\\
        &V_{v_i} = \begin{pmatrix}  w_{uf_i}^T+w_v^T & 0  & \cdots&0 \end{pmatrix}^T
\end{aligned}
\end{equation}
To estimate the unknown foot model noise covariance, we use the innovation-based covariance matching technique \cite{SONG2020AKF}. Each $i$-th foot innovation $e_i$ is given by :
\begin{equation}\small
\begin{split}
       e_i &= s_v (\chi^- Y_{v_i}-b_v)= s_v (\eta^R b_v -b_v + \chi^- V_{v_i}) \\
       & = H_v\xi^-+M_v(w_{uf_i}+w_v) \\
       \text{where, } \quad &s_v=\begin{pmatrix}
           I_{3,3} &0_{3,3}&\cdots&0_{3,3} 
       \end{pmatrix}\\
       &H_v=\begin{pmatrix}
       0_{3,3} &I_{3,3} &0_{3,3}&\cdots&0_{3,3}    
       \end{pmatrix}\\
       &M_v = R_B
       \end{split}
\end{equation}
The covariance of innovation is derived as follows:
\begin{equation}
\begin{split}
     \mathbb{E}[e_i e_i^T] &= H_v P^- H_v^T + R_B(Q_{uf_i}+Q_{v})R_B^T,\\
     \text{where,}\quad  Q_{uf_i}&=\mathbb{E}[w_{uf_i}w_{uf_i}^T], \quad Q_{v}=\mathbb{E}[w_v w_v^T]
\end{split}
\end{equation}
The predicted error covariance $P^-$ is predicted with foot model noise $Q_{f_i}=\mathbb{E}[w_{f_i}w_{f_i}^T]$. The covariance of innovation can be approximated using a moving window as follows:
\begin{equation}
    E[e_{i} e_{i}^T] \approx \frac{1}{m}\sum_{n=0}^{m-1}e_{i}(k-n)e_{i}(k-n)^T = U_i(k)
\end{equation}

In this case, a small window size (5-10) is used to respond quickly to slips. If the number of data points is less than the size of the moving window, the past innovation is assumed to be zero. The estimated unknown covariance of foot model noise $\hat{Q}_{uf_i}$ is derived as follows:
\begin{equation}\label{eq:covariance_matching}
    \hat{Q}_{uf_i} = R_B^T(U_i - H_v P^-_k H_v^T)R_B - Q_v
\end{equation}
Here, $Q_v$ also serves to adjust the zero point in the covariance estimation. Therefore, $Q_v$ can be tuned such that the $\hat{Q}_{uf_i}$ becomes close to zero in the static state. We assume that the $\hat{Q}_{uf_i}$ is $\alpha$ times the constant covariance $Q_{f_i}$. The scaling factor matrix $\alpha$ is diagonal. Its each diagonal element $\alpha_{jj}$ computed as the ratio of the corresponding diagonal elements of $\hat{Q}_{uf_i}$ and $Q_{f_i}$:
\begin{equation}\label{eq:alpha}
\begin{split}
     \alpha_{jj} &= \frac{\hat{Q}_{uf_i,jj}}{Q_{f_i,jj}} \\
     \alpha_{jj} &\leftarrow \min\left(\max( \alpha_{jj}, 1), \alpha_{\text{max}}\right)\quad \text{for} \quad j = x, y, z.
\end{split}
\end{equation}
To prevent filter divergence, the scaling factor matrix $\alpha$ is clipped to have maximum and minimum values. Adjust the foot noise covariance as:
\begin{equation}\label{eq:scaling_covariance}
    \hat{Q}_{f_i} \leftarrow \alpha Q_{f_i},
\end{equation}
The covariance prediction step \eqref{eq:p_prediction} is performed again with $\hat{Q}_{f_i}$.

\subsection{Adaptive Filtering - Slip Rejection}
Kalman filter computes optimal estimates by minimizing error covariance; therefore, it is vulnerable to outliers. Consequently, if a slip occurs or the contact detection is temporarily incorrect, it acts as a significant outlier. We apply slip rejection using the Mahalanobis distance, commonly used in VKM\cite{Teng2021LeggedRS}, to PKM. The Mahalanobis distance is computed by a static-assumed velocity kinematics equation.

\begin{subequations}
    \begin{align}
        S & = H_v P_k^-H_v^T+M_v Q_v M_v^T\\
        d_i&= e_i^T S^{-1} e_i
    \end{align}
\end{subequations}
If the Mahalanobis distance of each $i$-th foot, $d_i$, exceeds a certain threshold, $d_i>\sigma$, the corresponding foot noise is set large (similar to the swing phase), followed by redoing the covariance prediction step \eqref{eq:p_prediction}. 
we provide the whole algorithm flowchart in Fig \ref{fig:algorithm_flow_chart}.

\section{Experimental Validation}
The proposed method was implemented and evaluated on our quadruped robot, LeoQuad. The true state was measured using motion capture equipment, as shown in Fig. \ref{fig:real_robot}. We conducted experiments with various gaits and terrain for 60 seconds, comparing four filter methods: conventional IEKF, IEKF with slip rejection (SR), IEKF with Foot noise Estimation (FE), and IEKF with both SR and FE. Performance was analyzed by the Root Mean Square Error (RMSE) of observable states: velocity in the body frame ($^{B}v = R_B^Tv_B$) and orientation (roll, pitch). We set $\alpha_{max}=9$. 
\begin{figure}[htbp]
    \centering
    \includegraphics[angle=270, width=0.5\textwidth, ]{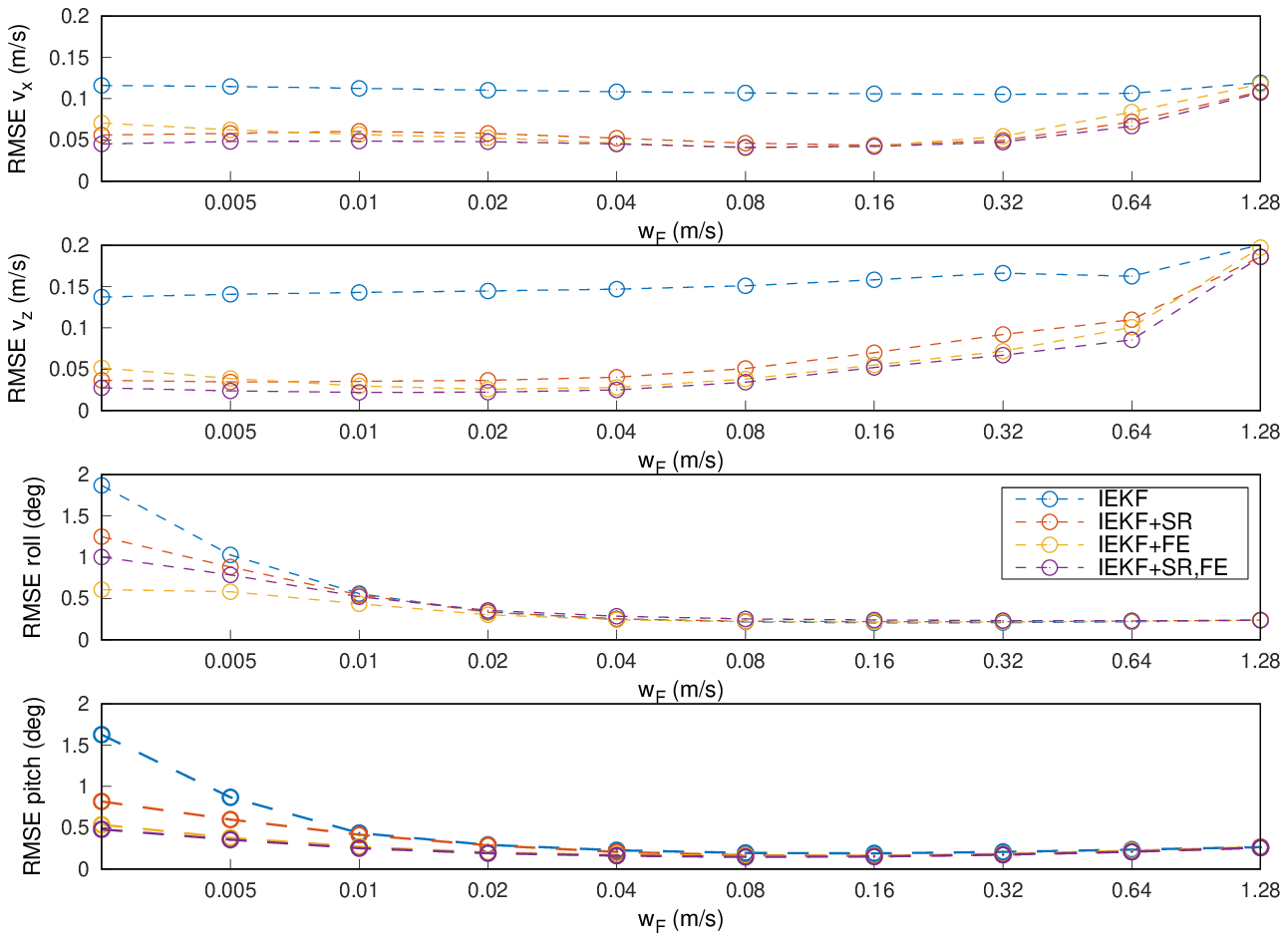}
    \caption{Graph showing the relationship between foot noise and the Root Mean Square Error(RMSE), illustrating the effect of noise levels on the filter accuracy. The experiments were conducted on rough terrain using a flying trot. 
    }
    \label{fig:flyingtrot_terrain} 
\end{figure}
\begin{figure}[htbp]
    \includegraphics[width=0.45\textwidth]{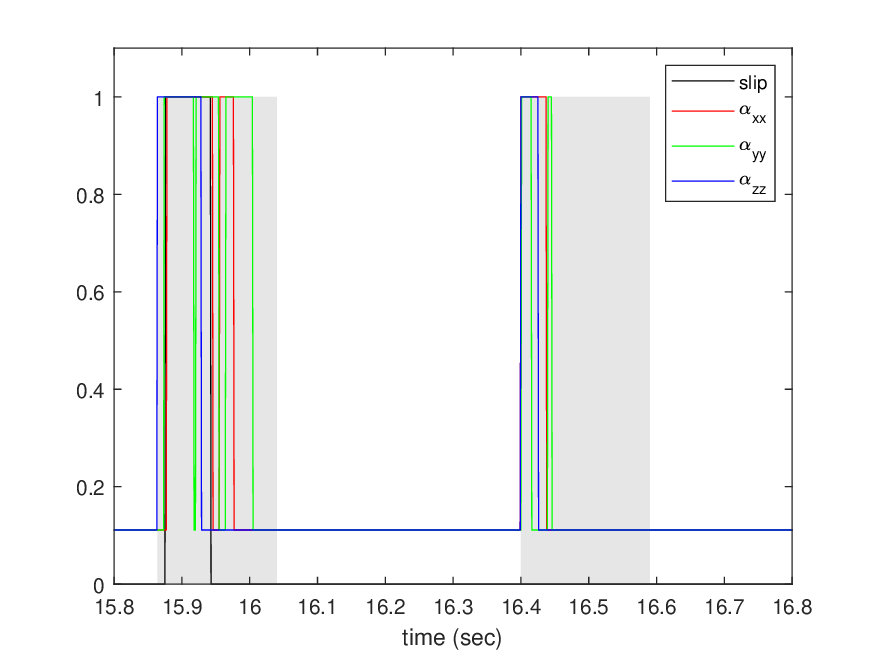} 
    \caption{Estimated contact, slip, and $\alpha$ values of the left front leg during the flying trot on rough terrain. $\alpha$ is normalize by $\alpha_{max}$ for readability. The gray-shaded regions indicate contact phases.}
    \label{fig:slip_and_alpha} 
\end{figure}

Fig. \ref{fig:flyingtrot_terrain} shows the performance of the flying trot on rough terrain under different foot noise $w_f$ settings. As the foot noise is set larger, the velocity error increases (especially in the z-axis), causing significant drift. Conversely, as the foot noise is set smaller, the orientation error increases. This highlights the difficulty of finding an optimal foot noise level for the conventional IEKF.

Fig. \ref{fig:slip_and_alpha} presents the contact detection, slip rejection, and normalized $\alpha$ values of the left front leg during the flying trot on rough terrain. The gray-shaded regions indicate contact. Around 15.9 seconds, slip rejection is observed, and the $\alpha$ responds first. Some $\alpha$ continue to increase even after the slip. Furthermore, after the contact impact, the $\alpha$ is not increased throughout the pre-swing phase. 

The performance of the filters under different gaits and terrains in $w_{f_i}=0.02$ is presented in Table \ref{tab:result}. 
For the trot gait on flat terrain, where the motion is relatively stable, all filters showed similar performance, with only minor differences between the conventional IEKF and the proposed methods. However, as the gait and terrain conditions became more dynamic, the IEKF with SR, IEKF with FE, and the combined SR+FE method exhibited significantly improved performance compared to the conventional IEKF. In the most dynamic scenario, flying trot in rough terrain, the IEKF with FE and the SR+FE method outperformed both the conventional IEKF and the IEKF with SR alone. While the conventional IEKF exhibited a substantial increase in velocity estimation error, the proposed methods maintained relatively low error growth, demonstrating their enhanced accuracy in high-dynamics conditions.

\begin{table}[htbp]
    \centering
    \setlength{\tabcolsep}{3pt}
    \caption{RMSE of velocity and orientation errors for different gaits and filters on flat and rough terrain $w_{f_i}=0.02$}
    \label{tab:result}
    \renewcommand{\arraystretch}{1.2}
    \begin{tabular}{c|c|c|ccc|cc}
        \hline
        Terrain & Gait & Filter & $^{B}v_x$ & $^{B}v_y$ & $^{B}v_z$ & Roll & Pitch \\
        & & & (m/s) & (m/s) & (m/s) & (deg) & (deg) \\
        \hline
        \multirow{12}{*}{Flat}  
              & \multirow{4}{*}{Trot} & IEKF     &\textbf{0.033}&0.022&\textbf{0.022}&0.330&\textbf{0.167}\\
              &                              & IEKF+SR  &0.036&0.022&0.028&0.331&0.179\\
              &                              & IEKF+FE  &0.037&\textbf{0.019}&0.032&\textbf{0.317}&0.177\\
              &                             & \textbf{IEKF+SR,FE}  &0.038&\textbf{0.019}&0.033&\textbf{0.317}&0.180 \\
        \cline{2-8}
                      & \multirow{4}{*}{Flying Trot} & IEKF      &0.075&0.021&0.073&0.162&\textbf{0.167}\\
              &                              & IEKF+SR  &0.037&\textbf{0.017}&0.044&\textbf{0.133}&0.216 \\
              &                              & IEKF+FE  &0.034&0.034&\textbf{0.043}&0.163&0.226\\
              &                             & \textbf{IEKF+SR,FE}   &\textbf{0.022}&0.028&0.045&0.192&0.236\\
        \cline{2-8}
              & \multirow{4}{*}{Pronk}       & IEKF      &0.029&0.045&0.063&0.195&0.222 \\
              &                              & IEKF+SR  &0.024&0.036&\textbf{0.049}&0.183&0.155  \\
              &                              & IEKF+FE   &\textbf{0.022}&\textbf{0.028}&0.055&\textbf{0.169}&0.136\\
               &                             & \textbf{IEKF+SR,FE} &0.023&\textbf{0.028}&0.054&0.198&\textbf{0.134} \\
        \hline
        \multirow{8}{*}{Rough} 
            & \multirow{4}{*}{Trot}   & IEKF     &0.037&0.030&0.032&0.352&0.249\\
            &                         & IEKF+SR  &0.032&0.029&0.026&\textbf{0.308}&0.233 \\
            &                         & IEKF+FE  &\textbf{0.026}&\textbf{0.028}&\textbf{0.027}&0.324&\textbf{0.226} \\
            &                         & \textbf{IEKF+SR,FE}   &0.027&0.029&0.028&0.320&0.232  \\
        \cline{2-8}
          & \multirow{4}{*}{Flying trot} & IEKF      &  0.110& 0.056 & 0.145 & 0.337 &  0.292\\
              &                          & IEKF+SR   &  0.058&  0.057&  0.037&  0.333& 0.286   \\
              &                          & IEKF+FE   & 0.053& \textbf{0.050}  &0.026 &\textbf{0.303} &0.199  \\
           &                             & \textbf{IEKF+SR,FE}   & \textbf{0.048} & \textbf{0.050} & \textbf{0.022}   & 0.356 & \textbf{0.190} \\
        \hline
    \end{tabular}
\end{table}

\section{Conclusion}


This paper introduced an Adaptive Invariant Extended Kalman Filter for proprioceptive state estimation in legged robots. Our method dynamically adjusts the contact foot model noise using online covariance estimation, improving estimation accuracy in dynamic conditions. In contrast to conventional approaches with fixed noise assumptions or slip rejection, our method enhances state estimation by adapting to slip conditions without requiring additional sensors.
We validated our method through real robot experiments without contact sensors. The results showed improved velocity and orientation estimation compared to the conventional IEKF and threshold-based slip rejection. 
Future work will refine tuning noise parameters and extend the method to other legged robots. It also includes improving the estimation of unobservable states like position and yaw.


%

\bibliographystyle{IEEEtran}
\bibliography{reference}

\end{document}